\documentclass{article}
\usepackage{spconf,amsmath,graphicx,hyperref}
\usepackage{booktabs}
\usepackage[table,xcdraw]{xcolor}

\usepackage[most]{tcolorbox} 
\newtcolorbox{promptbox}{
  colback=gray!5,    
  colframe=gray!60,  
  fonttitle=\bfseries,
  coltitle=black,
  boxrule=0.8pt,
  arc=3pt,           
  left=6pt, right=6pt, top=6pt, bottom=6pt,
  title=Prompt
}

\title{Benchmarking Multimodal Large Language Models\\ for Face Recognition}
\name{
	Hatef Otroshi Shahreza and S\'{e}bastien Marcel
	\thanks{This work was funded by the Hasler foundation through the Responsible Face Recognition (SAFER) project
    and also by the European Union project CarMen (Grant Agreement No. 101168325).
	}
}
\address{Idiap Research Institute, Switzerland\\
   {\{hatef.otroshi, sebastien.marcel\}@idiap.ch}
}%
\begin{document}
%
\maketitle

\begin{abstract}
Multimodal large language models (MLLMs) have achieved remarkable performance across diverse vision-and-language tasks. However, their potential in face recognition remains underexplored. In particular, the performance of open-source MLLMs needs to be evaluated and compared with existing face recognition models on standard benchmarks with similar protocol. In this work, we present a systematic benchmark of state-of-the-art MLLMs for face recognition on several face recognition datasets, including LFW, CALFW, CPLFW, CFP, AgeDB and RFW. Experimental results reveal that while MLLMs capture rich semantic cues useful for face-related tasks, they lag behind specialized models in high-precision recognition scenarios in zero-shot applications. This benchmark provides a foundation for advancing MLLM-based face recognition, offering insights for the design of next-generation models with higher accuracy and generalization. The source code of our benchmark is publicly available in the \href{https://www.idiap.ch/paper/facerecbench/}{project page}.
\end{abstract}

\section{Introduction}
\label{sec:intro}

Multimodal large language models (MLLMs) have recently gained significant attention from the research community for visual and linguistic understanding tasks. By combining pretrained visual encoders with large language models (LLMs), systems such as Flamingo~\cite{alayrac2022flamingo}, QwenVL~\cite{Qwen2VL}, and GPT-4o~\cite{hurst2024gpt} have achieved state-of-the-art performance across diverse tasks, including image captioning and visual question answering (VQA). These models showcase the ability of LLMs to reason over perceptual inputs and generate coherent, contextually grounded output text, enabling general-purpose image processing in zero-shot and few-shot settings. 
Leveraging large-scale pretraining, they have accelerated the development of foundation models that are capable of interpreting and responding to complex visual questions without requiring task-specific supervision.

Face recognition is also a popular computer vision and is increasingly used in different applications \cite{deng2019arcface,george2024edgeface,kim2022adaface}. In particular, face recognition is used as a secure authentication tool in a broad range of applications such as smart phone unlocking, border control, etc. In addition to the security purposes, face recognition is used for entertainment and also in social media. 
Face recognition models have been extensively studied in the literature and
there are also standard benchmarks to evaluate and compare the performance of face recognition models. 

With the surge of MLLMs, we can consider potential applications of MLLMs for face recognition~\cite{shahreza2025foundation}. However, for replacing MLLMs with existing face recognition models, it is important to know the performance of MLLMs compared to typical models on standard benchmarks with similar protocol. 
In this paper, we investigate \textit{how open-source MLLMs perform on face recognition benchmarks?} 
While there are some previous works on evaluation of MLLMs for different face understanding tasks~\cite{narayan2025facexbench,hassanpour2024chatgpt,komaty2025exploring}, to our knowledge this paper is the first work that benchmarks MLLMs for face recognition on standard datasets with similar protocols.

In the remaining of this paper, we first review previous work in the literature in Section~\ref{sec:related-work}. We then describe our benchmark in Section \ref{sec:benchmark} and present our results in Section \ref{sec:results}. Finally, the paper is concluded in Section \ref{sec:conclusion}.

\section{Related Work}\label{sec:related-work}
Recently, several papers have investigated the application of MLLMs  for face-related tasks, including face recognition, attribute analysis, forgery detection,  anti-spoofing, and multimodal reasoning. A recent survey \cite{shahreza2025foundation} provides an extensive
review of how foundation models and MLLMs are being applied in biometrics and face recognition.


\begin{figure*}[htbp]
    \centering
    \includegraphics[width=0.975\linewidth,trim={3cm 23.7cm 3cm 3.5cm},clip]{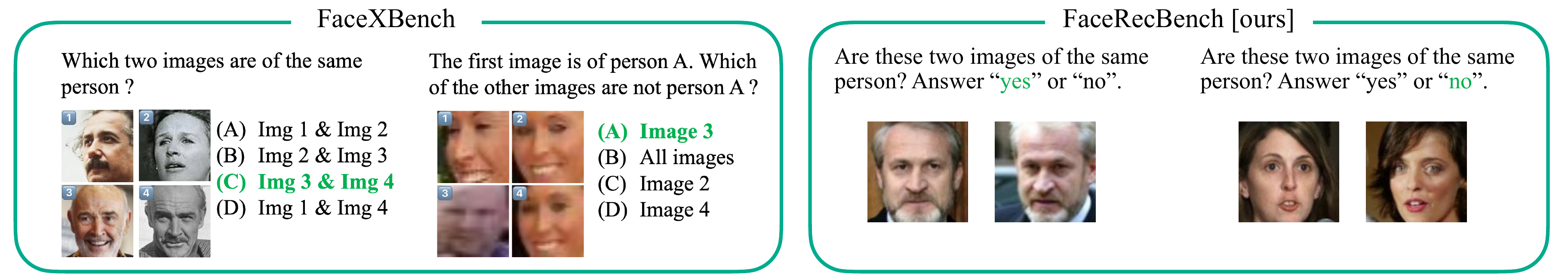}
    \caption{Sample questions in FaceXBench \cite{narayan2025facexbench} and our benchmark.}
    \label{fig:facexbench}
\end{figure*}

Early studies investigated the use of pretrained MLLMs, such as ChatGPT~\cite{hurst2024gpt}, for face verification~\cite{hassanpour2024chatgpt}, and predicting soft-biometrics, such as age, gender, and ethnicity. 
Jia \textit{et al.} \cite{jia2024can} also evaluated the application of ChatGPT for zero-shot face deepfake detection.  Shi \textit{et al.} \cite{shi2024shield} explored chain-of-thoughts prompting for  Gemini and ChatGPT in face anti-spoofing and deepfake detection tasks. 
Komaty \textit{et al.} ~\cite{komaty2025exploring} investigated in-context learning of ChatGPT~\cite{hurst2024gpt} for face anti-spoofing. Sony \textit{et al.} \cite{sony2025foundation} evaluated the performance of several foundation models (such as CLIP, BLIP, etc.) for face recognition, and showed that fusion of face recognition models with foundation models can improve recognition accuracy.

In addition to these studies, some benchmarks were proposed for different face processing tasks. FaceBench~\cite{wang2025facebench} proposed a visual question-answering benchmark for facial attributes. Benchmarks such as FaceXBench~\cite{narayan2025facexbench} and Face-Human-Bench~\cite{qin2025face} were also proposed to benchmark MLLMs across various face processing tasks, including facial expression recognition, attribute prediction,  anti-spoofing, etc. 
FaceXBench~\cite{narayan2025facexbench} also includes face recognition, using face recognition datasets such as 
LFW~\cite{huang2008labeled}, 
AgeDB~\cite{moschoglou2017agedb}, 
CFP-FF~\cite{sengupta2016frontal}, 
CFP-FP~\cite{sengupta2016frontal}, 
CALFW~\cite{zheng2017cross}, 
CPLFW~\cite{zheng2018cross}. However, they used multiple-choice questions for evaluating the performance of MLLMs. Fig.~\ref{fig:facexbench} illustrates two example questions from FaceXBench for face recognition task. While FaceXBench is a useful benchmark for comparing MLLMs in face processing tasks, the reported accuracy values are not comparable to the accuracy of face recognition models in the literature. In fact, for evaluating typical face recognition models, we simply have two face images and want to see if they have same identity. However, reported values for face recognition based on  questions with multiple images  and also multiple choices are not consistent with values reported in the literature. 
In this paper, we focus on face recognition and benchmark MLLMs with similar protocols as typical face recognition models.

\section{Benchmarking MLLMs for Face Recognition}\label{sec:benchmark}
To evaluate MLLMs for face recognition, we consider a verification task where two face images are available and the question is whether the given images belong to the same identity. Hence, we give the MLLM with both images along with the following prompt:
\begin{promptbox}
Are these two images of the same person? Answer ``yes" or ``no".
\end{promptbox}
Then, the output of MLLM is expected to be ``yes" or ``no", meaning if the images are predicted to correspond to same person or not.

In our benchmark, we consider different standard datasets, including Labeled Faces in the Wild (LFW) \cite{huang2008labeled}, Cross-age LFW (CALFW) \cite{zheng2017cross}, Cross-Pose LFW (CPLFW) \cite{zheng2018cross}, Celebrities in Frontal-Profile in the Wild (CFP) \cite{sengupta2016frontal}, AgeDB-30 \cite{moschoglou2017agedb}, and Racial Faces in-the-Wild (RFW) \cite{wang2019racial}. 
Our evaluation for each of these dataset include 6,000 pairs of images with 3,000 positive and 3,000 negative pairs. For consistency with prior works on face recognition, we report recognition accuracy on these datasets.  In the following, we briefly describe each dataset:

\begin{table*}[tbh]
    \centering
\setlength{\tabcolsep}{9pt}
        \caption{Comparison of recognition accuracy (\%) of MLLMs with face recognition models on different face datasets.}
    \label{tab:results}

\scalebox{0.95}{
\begin{tabular}{lccccccc}
\toprule
Model & LFW & AgeDB30 & CALFW & CPLFW & CFP-FP & CFP-FF & Average \\
\midrule
        \rowcolor[HTML]{eafaf1}
         \multicolumn{8}{c}{\textbf{Open source MLLMs}} 
        \\
LLaVA-v1.5-7b & 50.00 & 50.00 & 50.00 & 50.00 & 50.00 & 50.00 & 50.00 \\
LLaVA-v1.5-13b & 49.92 & 49.68 & 50.02 & 49.83 & 50.09 & 50.01 & 49.93 \\
LLaVA-OneVision-Qwen2-0.5b & 55.52 & 52.63 & 50.08 & 51.55 & 55.00 & 55.61 & 53.40 \\
LLaVA-NeXT-Vicuna-13b & 65.95 & 57.50 & 51.53 & 54.95 & 67.44 & 70.87 & 61.37 \\
LLaVA-NeXT-Vicuna-7b & 53.02 & 50.72 & 50.13 & 50.13 & 53.87 & 56.86 & 52.45 \\
LLaVA-NeXT-Mistral-7b & 50.00 & 49.98 & 50.02 & 50.22 & 49.99 & 50.03 & 50.04 \\
GLM-4v-9b & 52.58 & 50.42 & 48.52 & 50.27 & 50.39 & 49.93 & 50.35 \\
Idefics-9b-Instruct & 50.13 & 50.05 & 50.02 & 49.98 & 49.99 & 50.40 & 50.09 \\
Idefics2-8b & 72.40 & 68.53 & 54.93 & 55.98 & 74.11 & 74.43 & 66.73 \\
Idefics3-8B-Llama3 & 88.83 & 57.98 & 61.18 & 70.90 & 80.26 & 85.11 & 74.05 \\
ShareGPT4v-7b & 49.98 & 50.00 & 50.00 & 49.95 & 50.00 & 50.00 & 49.99 \\
ShareGPT4v-13b & 49.92 & 49.98 & 50.00 & 49.87 & 50.16 & 50.00 & 49.99 \\
PaliGemma-3b-mix-448 & 48.48 & 49.40 & 48.43 & 50.37 & 50.26 & 50.33 & 49.54 \\
Ovis1.5-Llama3-8B & 52.63 & 52.40 & 50.85 & 50.23 & 53.67 & 54.97 & 52.46 \\
Ovis1.5-Gemma2-9B & 73.93 & 57.87 & 56.95 & 54.93 & 75.09 & 78.49 & 66.21 \\
Llama-3.2-11B-Vision-Instruct & 50.55 & 48.20 & 51.83 & 49.50 & 49.47 & 49.94 & 49.92 \\
InternVL2.5-1B & 54.22 & 50.62 & 50.80 & 51.10 & 51.44 & 51.26 & 51.57 \\
InternVL3-1B & 69.28 & 56.25 & 56.15 & 63.35 & 60.80 & 64.06 & 61.65 \\
InternVL3-8B & 87.92 & 52.27 & 59.08 & 72.30 & 79.20 & 81.84 & 72.10 \\
InternVL3-38B & 90.10 & 55.72 & 61.37 & 71.20 & 72.50 & 84.40 & 72.55 \\
FaceLLM-8B & 90.65 & 53.38 & 61.48 & 73.50 & 80.06 & 84.89 & 73.99 \\
Valley2 & 92.93 & 60.75 & 68.58 & 74.55 & 84.33 & 92.27 & 78.90 \\
Qwen2-VL-2B-Instruct & 63.38 & 58.55 & 50.77 & 51.17 & 61.27 & 62.33 & 57.91 \\
Qwen2-VL-7B-Instruct & \textbf{93.28} & \textbf{66.03} & \textbf{71.95} & \textbf{75.28} & \textbf{86.93} & \textbf{93.11} & \textbf{81.10} \\
Qwen2.5-VL-3B-Instruct & 77.52 & 54.03 & 60.92 & 59.43 & 67.00 & 82.70 & 66.93 \\
Qwen2.5-VL-7B-Instruct & 89.48 & 59.07 & 69.08 & 73.43 & 79.09 & 89.46 & 76.60 \\
Qwen2.5-VL-32B-Instruct & 79.32 & 59.07 & 64.48 & 66.15 & 69.70 & 83.01 & 70.29 \\
\midrule
        \rowcolor[HTML]{FFF7ED}
         \multicolumn{8}{c}{\textbf{Face Recognition Models}} 
        \\
IResNet-50 (HyperFace) & 98.27 & 90.40 & 91.48 & 85.60 & 92.24 & 98.86 &  92.81 \\
IResNet-50 (MS1MV2) & \textbf{99.83} & \textbf{98.28} & \textbf{95.45} &  \textbf{92.08}  & \textbf{98.27} & \textbf{99.99} & \textbf{97.31} \\
\bottomrule
\end{tabular}
    }
\vspace{-5pt}    
\end{table*}

\textbf{Labeled Faces in the Wild (LFW):}
LFW~\cite{huang2008labeled} is a widely used benchmark dataset for unconstrained face verification. It contains 13,233 images of faces collected from the web, with 5,749 unique individuals. The dataset is designed to evaluate how well face recognition algorithms generalize to real-world conditions, with variations in pose, expression, illumination, and background. Since its release, LFW has served as a standard reference point for measuring progress in face recognition under unconstrained settings.

\textbf{Cross-Age LFW (CALFW):}
CALFW~\cite{zheng2017cross} extends LFW by introducing cross-age variation, aiming to make the verification task more challenging. It contains image pairs of the same individual captured at different ages, highlighting the difficulty of recognizing faces over long time spans. This dataset is primarily focused on age-related intra-class variations while maintaining inter-class diversity, making it a valuable benchmark for studying the robustness of face recognition systems to aging effects.

\textbf{Cross-Pose LFW (CPLFW):}
CPLFW~\cite{zheng2018cross} is another variant of LFW, created to evaluate face recognition under cross-pose conditions. It includes face pairs where the same subject appears in significantly different poses, thus introducing large intra-class variations in pose. By emphasizing pose differences, CPLFW complements LFW and CALFW to how well recognition systems handle extreme viewpoint changes.

\textbf{Celebrities in Frontal-Profile (CFP):}
The CFP~\cite{sengupta2016frontal} dataset is introduced to test face recognition across frontal and profile views. It consists of images of 500 celebrities, with both frontal and profile face shots, and provides verification protocols for frontal-to-frontal (CFP-FF) and frontal-to-profile (CFP-FP) matching.

\textbf{AgeDB:}
AgeDB \cite{moschoglou2017agedb}  is a benchmark dataset focused on age-related variations in face recognition. It contains images 16,516 images of 570 subjects with a wide age range. The dataset provides predefined verification protocols with increasing age gaps (e.g., 5, 10, 20, and 30 years), enabling systematic evaluation of how well algorithms handle the challenge of age progression. AgeDB is commonly used to study long-term face recognition performance. We use 30-year protocol in our benchmark.

\textbf{Racial Faces in-the-Wild (RFW):}
The RFW \cite{wang2019racial} dataset was introduced to evaluate bias and fairness in face recognition systems across different demographic groups. RFW is constructed by reorganizing images from MS-Celeb~\cite{guo2016ms} into four balanced subsets: Caucasian, Asian, Indian, and African. Each subset contains approximately 10,000 images from around 3,000 individuals, with 6,000 comparisons. By providing a benchmark focused on racial diversity, RFW enables systematic analysis of demographic disparities in recognition performance and has become a widely used dataset for studying fairness in face recognition.

We use the cropped images for each dataset available in Insightface repository\footnote{https://github.com/deepinsight/insightface}. We also use VLMEvalKit repository\footnote{https://github.com/open-compass/VLMEvalKit} to implement our benchmark.
The source code of our benchmark is publicly available in the \href{https://www.idiap.ch/paper/facerecbench/}{project page}\footnote{Project page: \href{https://www.idiap.ch/paper/facerecbench/}{https://www.idiap.ch/paper/facerecbench}}.

\begin{table}[tbh]
    \centering

\setlength{\tabcolsep}{2.25pt}
        \caption{Comparison of recognition accuracy (\%) of MLLMs on different demography groups in RFW.}
    \label{tab:results_rfw}

\scalebox{0.78}{
\begin{tabular}{lcccccc}
\toprule
Model & African & Asian & Caucasian & Indian & Average & Std. \\
\midrule
        \rowcolor[HTML]{eafaf1}
         \multicolumn{7}{c}{\textbf{Open source MLLMs}} 
\\
LLaVA-NeXT-Vicuna-13b & 51.28 & 53.53 & 56.88 & 53.00 & 53.67 & 2.03 \\
Idefics3-8B-Llama3 & 60.78 & 66.15 & 70.38 & 66.38 & 65.92 & 3.41 \\
Ovis1.5-Gemma2-9B & 52.62 & 54.92 & 61.93 & 55.83 & 56.33 & 3.44 \\
InternVL3-8B & 64.37 & 63.08 & 66.37 & 64.03 & 64.46 & 1.20 \\
FaceLLM-8B & \textbf{66.00} & 66.78 & 68.82 & 66.02 & 66.90 & \textbf{1.15 }\\
Valley2 & 63.53 & \textbf{68.77} & 75.57 & \textbf{70.28} & \textbf{69.54} & 4.29 \\
Qwen2-VL-7B-Instruct & 60.40 & 65.55 & \textbf{76.68} & 68.35 & 67.75 & 5.90 \\
Qwen2.5-VL-7B-Instruct & 62.65 & 66.63 & 70.55 & 68.98 & 67.20 & 2.98 \\
\midrule
\rowcolor[HTML]{FFF7ED}
         \multicolumn{7}{c}{\textbf{Face Recognition Models}} 
        \\
IResNet-50 (HyperFace) & 88.27 & 82.98 & 84.33 & 78.02 & 83.40 & 3.66 \\
IResNet-50 (MS1MV2) & \textbf{98.32} & \textbf{97.73} & \textbf{99.33} & \textbf{98.23} & \textbf{98.40} & \textbf{0.58} \\
\bottomrule
\end{tabular}
}
\vspace{-5pt}    
\end{table}

\section{Experimental Results}\label{sec:results}
We evaluate and benchmark open-source\footnote{Note that given the restrictions in license of each of the benchmark datasets, we were not able to use commercial MLLMs in this study.} MLLMs on various standard face recognition datasets, including LFW \cite{huang2008labeled}, CALFW \cite{zheng2017cross}, CPLFW \cite{zheng2018cross}, CFP \cite{sengupta2016frontal}, AgeDB-30 \cite{moschoglou2017agedb}, and RFW \cite{wang2019racial}. The MLLMs used in our experiments include
LLaVA-v1.5-7b~\cite{liu2023llava}, 
LLaVA-v1.5-13b~\cite{liu2023llava}, 
LLaVA-OneVision-Qwen2-0.5b~\cite{liu2023improvedllava}, 
LLaVA-NeXT-Vicuna-7b~\cite{liu2023improvedllava}, 
LLaVA-NeXT-Vicuna-13b~\cite{liu2023improvedllava}, 
LLaVA-NeXT-Mistral-7b~\cite{liu2023improvedllava}, 
GLM-4v-9b~\cite{glm2024chatglm}, 
Idefics-2-8b~\cite{laurençon2024matters}, 
Idefics-9b-Instruct~\cite{laurencon2023obelics}, 
Idefics3-8B-Llama3~\cite{laurençon2024building}, 
ShareGPT4v-13b~\cite{chen2024sharegpt4v}, 
ShareGPT4v-7b~\cite{chen2024sharegpt4v}, 
PaliGemma-3b-mix-448~\cite{beyer2024paligemma}, 
Ovis-1.5-Llama3-8B~\cite{lu2024ovis}, 
Ovis1.5-Gemma2-9B~\cite{lu2024ovis}, 
Llama-3.2-11B-Vision-Instruct \cite{grattafiori2024llama}, 
InternVL2.5-1B~\cite{chen2024expanding}, 
Intern-VL3-1B~\cite{zhu2025internvl3}, 
InternVL3-2B~\cite{zhu2025internvl3}, 
InternVL3-8B~\cite{zhu2025internvl3}, 
Face-LLM-8B~\cite{shahreza2025facellm}, 
Valley2~\cite{wu2025valley2}, 
Qwen2-VL-2B-Instruct~\cite{Qwen2VL}, 
Qwen2-VL-7B-Instruct~\cite{Qwen2VL}, 
Qwen2.5-VL-3B-Instruct~\cite{qwen2025qwen25technicalreport}, 
Qwen2.5-VL-7B-Instruct \cite{qwen2025qwen25technicalreport}, 
Qwen2.5-VL-32B-Instruct \cite{qwen2025qwen25technicalreport}.
We run our evaluations on a system equipped with NVIDIA H100. 

Table \ref{tab:results} reports the performance of different MLLMs on several face recognition benchmarks (LFW, CALFW, CPLFW, CFP, and AgeDB-30). This table also compares the performance of MLLMs with IResNet-50 (trained with AdaFace loss \cite{kim2022adaface} on MS-Celeb~\cite{guo2016ms} dataset) as a state-of-the-art face recognition model. As another baseline, we also consider a face recognition model with IResNet-50 trained with HyperFace~\cite{shahrezahyperface} synthetic dataset. As the results in this paper show there is a significant  gap between the performance of face recognition models. While increasing the size of MLLM can improve the performance on benchmarks, it also saturates for each MLLM (can be seen in InternVL3 and Qwen2.5VL). 

Among the benchmarked MLLMs, FaceLLM is based on InternVL3 and finetuned for face understanding. The results in Table \ref{tab:results} show that the finetuing in FaceLLM has increased the performance compared to the base model (InternVL3) on different face recognition benchmarks. This suggests that by using domain-specific data instead of general-purpose data, we can expect improvement in MLLMs for the face recognition task.

We also compare the top-performing models in Table~\ref{tab:results} on RFW dataset. Table~\ref{tab:results_rfw} reports the performance of different models for four demography groups. The result in this table also indicate significant gap between the performance of MLLMs with typical models.

\section{Conclusion}\label{sec:conclusion}
In this paper, we presented a benchmark for MLLMs with similar protocol used to evaluate typical face recognition models.
Although MLLMs have shown considerable potentials in broad applications, most are trained mainly on general-purpose datasets or large-scale image–text pairs collected from the web. Consequently, these models are able to generate high-level image descriptions but often lack task-specific precision. For instance, while they can describe a person's appearance or identify basic demographic attributes such as age and gender, they frequently struggle with more details which are required to recognize identity or verify if identity is the same in two images. This limitation poses challenges for applications of MLLMs in face recognition and requires further study in future. 
Our benchmark can be used by future researchers to compare the MLLMs with face recognition models.

{
\small
\bibliographystyle{IEEEbib}
\bibliography{refs}
}
\end{document}